\documentclass[10pt,twocolumn,letterpaper]{article}

\usepackage{cvpr}
\usepackage{times}
\usepackage{epsfig}
\usepackage{graphicx}
\usepackage{amsmath}
\usepackage{amssymb}
\usepackage{subfig}
\usepackage{multirow}

\usepackage[pagebackref=true,breaklinks=true,letterpaper=true,colorlinks,bookmarks=false]{hyperref}
\cvprfinalcopy 

\def\cvprPaperID{219} 

\ifcvprfinal\fi
\begin{document}

\title{Discovering Underlying Person Structure Pattern with Relative Local Distance for Person Re-identification}

\author{Guangcong Wang\\
\and
Jianhuang Lai\\
\and
Zhenyu Xie\\
\and
Xiaohua  Xie\\
{\tt\small $\{$wanggc3,xiezhy6$\}$@mail2.sysu.edu.cn,}
{\tt\small $\{$stsljh,xiexiaoh6$\}$@mail.sysu.edu.cn}
}
\maketitle

\begin{abstract}
  Modeling the underlying person structure for person re-identification (re-ID) is difficult due to diverse deformable poses, changeable camera views and imperfect person detectors. How to exploit underlying person structure information without extra annotations to improve the performance of person re-ID remains largely unexplored. To address this problem, we propose a novel Relative Local Distance (RLD) method that integrates a relative local distance constraint into convolutional neural networks (CNNs) in an end-to-end way. It is the first time that the relative local constraint is proposed to guide the global feature representation learning. Specially, a relative local distance matrix is computed by using feature maps and then regarded as a regularizer to guide CNNs to learn a structure-aware feature representation. With the discovered underlying person structure, the RLD method builds a bridge between the global and local feature representation and thus improves the capacity of feature representation for person re-ID. Furthermore, RLD also significantly accelerates deep network training compared with conventional methods. The experimental results show the effectiveness of RLD on the CUHK03, Market-1501, and DukeMTMC-reID datasets. Code is available at \url{https://github.com/Wanggcong/RLD_codes}.
\end{abstract}

\section{Introduction}
\label{sec:intro}
\begin{figure}
  \centering
  \includegraphics[width=0.45\textwidth]{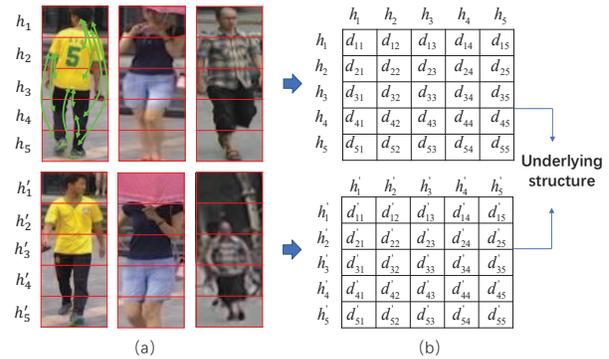}\\
  \caption{Motivation. (a) Examples of deformable person structure. Each column denotes one person ID. Three image pairs show that the person structure may be affected by pose/view changes, partial body detections, loose bounding box detections. (b) Proposed relative local distance matrix (computed by pair-wise patches, denoted as green allows in (a)). Even though the person structure is deformable, the relative local distances of person patches are quite robust, i.e.,  a person image consists of head, upper body, lower body and foot from top to bottom. The relative local distances are learnable and can discover the underlying person structure, which builds a bridge between the global and local feature and thus improves the performance of re-ID (Best viewed in color).}\label{fig:motivation}
\end{figure}

Person re-identification (ReID) aims to re-target pedestrian images across non-overlapped cameras given a probe image. Currently, most of person re-ID models \cite{} achieve a significant performance by using deep learning for feature representation. However, it is still an extremely challenging task because a person's appearance often undergoes large intra-class variations, e.g., significant changes in illumination, background clutter, pose, viewpoint, and occlusion. Among these variations, the deformable pose is one of the most challenging problems because feature representations of person images are very sensitive to the complex deformable poses (Figure \ref{fig:motivation} (a)) . How to exploit the person structure information to improve the performance of person re-ID remains an open problem.

Recent research reveals that lots of pose-driven methods \cite{Zhao_2017_CVPR,Su2017ICCV,sarfraz2017pose,Liu_2018_CVPR,wei2017glad}, which exploit human pose annotations from extra datasets, are greatly beneficial to improve the accuracy of person re-ID. These methods first train good pose estimators on human pose datasets and then predict the pose keypoints on re-ID datasets. With predicted pose keypoints, they can jointly learn both the global full-body and discriminative local body-part features for re-ID systems. However, these methods need huge amounts of expensive human pose annotations, which are limited in real-world scenarios.

Motivated by this point, a question arises: can we directly model the person structure without any extra annotation? Obstacles to answering this question are the large pose deformations and the complex view variations. Existing re-ID methods that may offer a partial solution to this problem can be generally categorized into two groups.

In the first group, researchers attempt to exploit the person structure information by simply assuming that person images are well aligned. With this strong assumption, they directly divided the person images into horizontal stripes or grids \cite{cheng2016person,yi2014deep,zheng2013reidentification} for local feature representations, which are further concatenated into one feature representation for global metric learning. However, such an assumption does not hold in many real-world scenarios due to diverse poses, different camera views, and inevitable detection errors, as shown in Figure \ref{fig:motivation} (a). In these scenarios, simply extracting features from stripes or grids of person images leads to the dramatic reduction of the performance.

In the second group, instead of using the strong assumption of person part alignment, some studies \cite{li2014deepreid,bkak2017deep,shen2015person} seek an effective approach to automatically align the person body part for local metric learning. For each local patch of a probe image, they find the most similar patch of a gallery image. The similarity between two images depends on  all of the corresponding patch pairs. However, the feature representation of local matching is dynamic and thus costs quite expensive computation when the gallery set is large. Suppose there are $M$ probe images and $N$ gallery images, patch matching methods need $M\times N$ on-line feature extraction operations while conventional matching methods need $M+N$ operations ($N$ operations can be off-line). Besides, it is hard to integrate local matching into CNNs because such an assignment problem (e.g., Hungary algorithm) is non-differentiable and thus cannot be integrated into CNNs using gradient descent methods for an end-to-end optimization. These two drawbacks limit their applications in real-world scenarios.

Different from these two kinds of re-ID methods, we propose a novel Relative Local Distance (RLD) method that integrates a person structure constraint into CNNs in an end-to-end way. Specially, RLD extends conventional classification networks by adding a small branch for exploiting person structure patterns. To achieve this, a relative local distance matrix is computed between different column feature vectors of feature maps and used to represent the structure of a person. After several fully connected layers,  a classification loss is used to learn a structure-aware feature representation for each person identity. With the discovered underlying person structure, RLD builds a bridge between the global and local feature and thus improves the capacity of feature representation for person re-ID. In addition, RLD significantly accelerates the deep network training by adding only a small overhead to conventional re-ID systems. Note that, RLD differs a lot from the second-group methods because local distances of RLD are computed between different patches within one image while the second-group methods perform patch matching between image pairs. As far as we know, it is the first time the relative local distance is proposed to exploit object structure for facilitating feature learning.

Compared with conventional re-ID methods, the proposed RLD method is featured in four aspects. First, RLD improves the performance of person re-ID without any pose annotation. Second, the person structure branch is a very small overhead, e.g., 0.8\% of the entire CNNs. Third, RLD significantly accelerates the deep network training. Fourth, the person structure of RLD is learnable and there is no strong assumption. Therefore, it can be also easily generalized to other object recognition problems.

Overall, this paper makes three main contributions.
\begin{itemize}
  \item First, it is the first time the relative local distance (RLD) is proposed to exploit the underlying person structure and facilitate feature learning without any pose annotation for person re-ID systems.
  \item Second, with the joint training of the person identity loss and person structure loss, RLD significantly accelerates deep network training compared with conventional methods and improves the performance of person re-ID with a very small overhead.
  \item Third, the experimental results show the effectiveness of RLD on the CUHK03, Market-1501, DukeMTMC-reID datasets.
\end{itemize}

The rest of the paper is organized as follows. Section \ref{sec:related} reviews the related work. Section \ref{sec:method} introduces our relative local distance model. The experimental comparisons, ablation studies, and analyses are presented in Section \ref{sec:expe}. Section \ref{sec:conc} concludes the paper.

\section{Related Work}
\label{sec:related}
Person re-identification (re-ID) \cite{zheng2013reidentification,zheng2015scalable,zhang2016learning,liao2015person} has been studied extensively in the past ten years. Currently, deep models, which adopt the identity loss \cite{zheng2016mars,wang2018st_ReID,FengLX18}, verification loss \cite{li2014deepreid,ahmed2015improved,chen2015deep,Farenzena2010Person}, triplet loss \cite{ding2015deep,wang2017tcsvt,HermansBL17,cheng2016person} or other metric functions \cite{lin2017cross,Wang2016DARI,Chen_2018_CVPR}, are successfully employed in person re-ID and achieve promising results due to the remarkable representation ability of CNNs.

Recently, researchers \cite{Zhao_2017_CVPR,Su2017ICCV,sarfraz2017pose,Liu_2018_CVPR,wei2017glad} attempted to exploit person structure information to learn local discriminative features. To this they trained human pose estimators on pose annotated datasets and used the pose estimators to help the feature learning of person re-ID. For example, a spindle network structure \cite{Zhao_2017_CVPR} is proposed to use human landmark annotations for body joint localization and body region generation, which further guided multi-stage feature decomposition and tree-structured competitive feature fusion. A pose-driven CNN model \cite{Su2017ICCV} leveraged human pose information and transformed a global body image into an image containing normalized part regions for feature learning. In addition, Sarfraz et. al. \cite{sarfraz2017pose} incorporated both the fine and coarse pose information of the person to learn a discriminative embedding by explicitly including this information into the learning process of a re-ID model. However, these methods depend on huge amounts of expensive human pose annotations.

Instead of using pose annotated datasets to train pose estimators, lots of approaches directly split person structure into stripes or grids \cite{cheng2016person,yi2014deep,zheng2013reidentification} for local feature representations that can be concatenated into one feature vector for further global metric learning. In \cite{yi2014deep}, person images are cropped into three overlapped stripes which are used to train three independent networks. At the score level, three networks are fused for metric learning. Similarly, \cite{cheng2016person} split a person image into three stripes and jointly learned both the global full-body and local body-parts features to improve the performance of person re-ID. These models assumed the person bodies are well aligned, which may fail in real-world scenarios.

Differently, some studies \cite{shen2015person,bkak2017deep,li2014deepreid} attempted to conduct local metric learning to align the person structure. For example, Shen et. al. \cite{shen2015person} introduced a boosting-based approach to learn a correspondence structure which indicates the patch-wise matching probabilities between images from a target camera pair. In \cite{bkak2017deep}, a patch-based deformable model is proposed to combine appearance term with a deformation cost that controls relative placement of patches. Li et. al. \cite{li2014deepreid} proposed a filter pairing neural network to handle misalignment and geometric transforms by integrating a patch matching layer into the neural network. Our RLD method differs a lot from these patch-based matching methods in that RLD focuses on self-structure exploration while patch-based matching methods focus on the matching between image pairs. Patch-based matching methods are not off-line algorithms and cannot extract gallery features in advance. When the gallery image set is very large, it costs too much time to extract deep features for person retrieval.

\section{Method}
\label{sec:method}
\begin{figure*}
  \centering
  \includegraphics[width=0.9\textwidth]{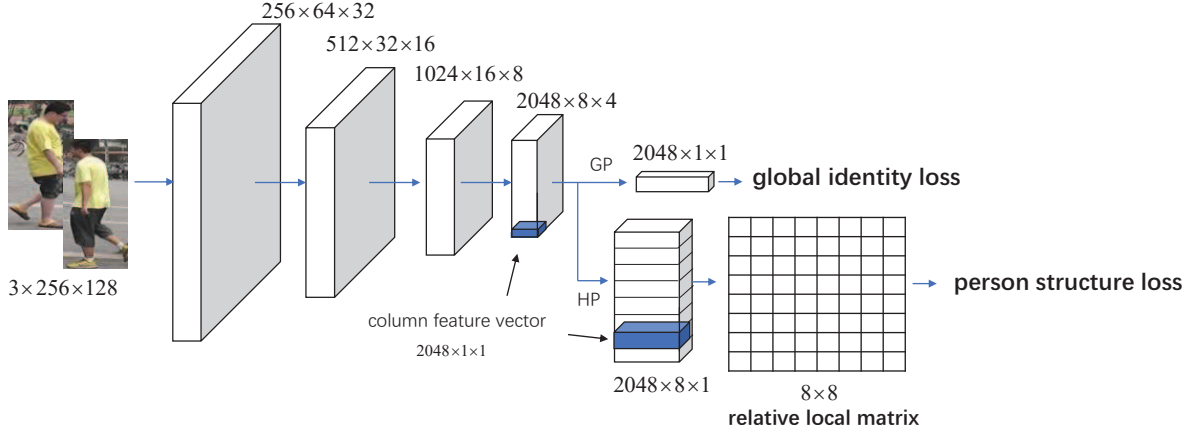}\\
  \caption{Overview of the relative local distance model. RLD uses ResNet50 as the convolutional backbone architecture, containing 4 building blocks. After the backbone architecture, there are two branches, i.e., global identity branch and person structure branch. The global identity branch consists of a 512-dim FC layer, a Batch Normalization layer, a ReLU layer, a dropout layer, a $K$-dim FC layer ($K$ denotes the number of person identities) and a Cross Entropy layer. Person structure branch also contains similar layers. The only difference is that the first dimension of Person structure branch is a 256-dimension. GP denotes a global average pooling operation and HP denotes a horizonal pooling operation (Best viewed in color).}\label{fig:overview}
\end{figure*}
The proposed Relative Local Distance (RLD) aims to exploit the underlying spatial structure of an object. Currently, a common approach of state-of-the-art Convolutional Neural Networks (CNNs) is to add a global average pooling (GAP) layer at the top of neural networks to avoid the over-fitting problem. However, this strategy partially drops the spatial structure information. In Section \ref{subsec:motivation}, we demonstrate the reason why GAP partially drops the structure information. To compensate for this, we introduce a novel relative local distance to describe the person structure in Section \ref{subsec:RLD} and \ref{subsec:learnable}. In Section \ref{subsec:proposed} and \ref{subsec:detail}, the proposed network architecture and implementation details are discussed, respectively.

\subsection{Drawbacks of Global Average Pooling.}
\label{subsec:motivation}
Let $I$ represent a RGB image of size $(3,h_{in},w_{in})$, $CNN$ represent a convolutional neural network. The output feature representation $f_{out}$ of size $(c_{out},h_{out},w_{out})$ is computed by $f_{out}=CNN(I)$. We define the concept of column feature vector as $c_{out}\times 1\times 1$ vector of $f_{out}$.  $f_{out}$ contains $h_{out}\times w_{out}$ column feature vectors. After $f_{out}$, several fully connected layers are directly used in some famous network architectures, e.g., AlexNet and VGG. To avoid over-fitting, an NIN network \cite{lin2013network} is proposed to replace the traditional fully connected layers by global average pooling. After that, the identity (classification) loss is computed by
\begin{equation}
    \label{eq:identity}
    {L_{id}} =  - \sum\limits_{k = 1}^K {{y_k}\log {p_k}}
\end{equation}
where $y_k$ is the $k$-th dimension value of the one-hot label $y$, $p_k$ represents the probability of the $k$-th person identity. However, directly pooling $(c_{out},h_{out},w_{out})$ feature maps into a feature vector $c_{out}\times 1 \times 1$ may discard some discriminative local features and lead to the confusion of object structure, especially fine-grained recognition tasks like person re-ID. For example, there are two persons with a similar appearance. The subtle differences are the type of shoes, shown in Figure \ref{fig:overview}. Without person structure constraints of $f_{out}$, GAP may partially drop subtle differences of person body parts by the average operation. As some empirical evidence of information loss in GAP, our observation is two-fold.



\emph{Observation 1:  spatial information of $f_{out}$ is available.} When we compute a spatial feature map of size $(h_{out}, w_{out})$ by averaging $c_{out}$ channels and use the spatial feature map as feature representation for person retrieval. Without the help of column feature vectors, we obtain 1.9\% rank-1 accuracy of performance on the CUHK03 dataset (compared with $\sim$0\% random ranking). This implies that $h_{out}\times w_{out}$ spatial feature maps contain available discriminative information and can be further exploited. $c_{out}$-dimension column feature vectors cannot fully encode all the spatial information. Directly using GAP to conduct an average pooling operation on all of column feature vectors leads to the spatial information loss.

\emph{Observation 2: column feature vectors are diverse.} When we compute relative local distances between different column feature vectors on  $f_{out}$ using the GAP trained model. We observe that neighbor column feature vectors are similar while distant column feature vectors are very different. This demonstrates that $f_{out}$ contains different kinds of discriminative feature representations for different image regions. It is not the best way to simply conduct the average pooling for CNNs when we want to capture the fine-grained discriminative regions.

From these two observations, we believe that although GAP avoids the over-fitting problem to some extent, it also partially drops spatial information. Considering the drawbacks of GAP, we are encouraged to develop an effective method to capture discriminative spatial information, \eg, the person structure information.

\subsection{Relative Local Distance}
\label{subsec:RLD}
Although there are diverse deformable person structures, an underlying object skeleton exists. For example, a person image consists of head, upper body, lower body and foot from top to bottom. To address the structure-confusion problem of GAP, an intuitive approach is to introduce a structure constraint for person re-ID. In this paper, we propose a learnable person structure by introducing a relative local distance without using pose annotations. Inspired by Observation 2, we model an underlying person structure using relative local distance comparison.

Given the output feature maps $f_{out}$ of size $(c_{out},h_{out},w_{out})$, we first use horizonal pooling to deal with viewpoint changes. We obtain vertical feature maps $f_v$ of size $(c_{out},h_{out},1)$. This is reasonable because different camera views may lead to dramatic horizonal changes of person structure \cite{liao2015person}. We then explore robust vertical person structures to guide the global feature representation learning. Specially, we split $f_v$ into $h_{out}$ column feature vectors with $c_{out}$ dimensions, denoted as ${\bf{x}}_i$, where $i=1,2,...,h_{out}$. We define a relative local distance $d_{ij}$ between two column feature vectors by computing the cosine distance
\begin{equation}
    \label{eq:d_ij}
    {d_{ij}} = {\bf{x}}_i^T{{\bf{x}}_j}
\end{equation}
where the relative local distance $d_{ij}$ builds a relationship between two local features. To simplify computation, we re-write Eq. \ref{eq:d_ij} as a matrix form
\begin{equation}
    \label{eq:D}
    {D} = {\bf{X}}^T{{\bf{X}}}
\end{equation}
where $D$ is a $h_{out}\times h_{out}$ relative distance matrix, describing similarities between different person body parts.



\subsection{Learnable Person Structure}
\label{subsec:learnable}
Mathematically, relative distance matrix $D$ is actually a complete structured probabilistic graph $G$ that we can model all kinds of interactions between all of random variables. Suppose a body part $x_i$ is a random variable (node), $d_{ij}$ is interaction (edge) between $x_i$ and $x_j$ random variables. A graph $G_k$ denotes a distribution over $h_{out}$-dimensional space, describing the dependencies of body parts for the $k$-th person. We use $G_1$, $G_2$,...,$G_K$ graphs to describe $K$ person identities. As a simple example, suppose a person $P_1$ wears white shirts and blue pants while another person $P_2$ wears white shirts and black pants. Let $h_{out}=2$, we obtain
\begin{equation}
\label{eq:G_ij}
{G_1} = \left[ {\begin{array}{*{20}{c}}
{1.0}&{0.2}\\
{0.2}&{1.0}
\end{array}} \right],{G_2} = \left[ {\begin{array}{*{20}{c}}
1.0&{0.0}\\
{0.0}&1.0
\end{array}} \right]
\end{equation}
where $0.2$ is the similarity score between white shirts and blue pants, and $0.0$  is the similarity score between white shirts and black pants. In this way, we can also distinguish different person appearances by using relative local distance matrixes that differ a lot from the absolute appearance feature representation.

$G_k$ is robust to diverse deformable person structures due to the fact that person bodies are structured in the order along the vertical axis. Affected by cameras views, deformable poses and imperfect detectors, the same body parts of a person collected from different camera views differ a lot in location, size and shape. To this we reshape a matrix $G_k$ into a feature vector and add two fully connected layers and a person structure loss to make the person structure learnable. By doing this, we can learn a local structure-aware feature representation to help the the global feature representation. The person structure loss in our network is the softmax loss,
\begin{equation}
    \label{eq:structure}
    {L_{stru}} =  - \sum\limits_{k = 1}^K {{y_k}\log {q_k}}
\end{equation}
where $y_k$ is the $k$-dimension value of the one-hot label $y$, $q_k$ represents the probability of the $k$-th person structure.

Finally, according to Eq. \ref{eq:identity} and Eq. \ref{eq:structure}, we define a structure-aware identity loss by
\begin{equation}
    \label{eq:loss}
    L = {L_{id}} + \lambda {L_{stru}}
\end{equation}
where $\lambda$ controls the relative importance of the identity loss and structure-aware loss.

\subsection{Network Architecture}
\label{subsec:proposed}
RLD is conceptually simple. Conventional classification networks contain one output for person identity prediction. RLD extends this pipeline and add a branch with a small overhead to guide neural networks be aware of the underlying person structure and then learns a structure-aware feature representation. The overview of our RLD framework is shown in Figure \ref{fig:overview}.

To demonstrate the generality of RLD, we use different architectures, ResNet18, ResNet34, ResNet50, and ResNet101. To simplify discussion, we take ResNet50 as an example. The other architectures will be discussed in Section \ref{sec:expe}. In Figure \ref{fig:overview}, RLD uses ResNet50 as the convolutional backbone architecture, containing 4 building blocks. After the backbone architecture, two branches are used, i.e., a global identity branch and a person structure branch. The global identity branch consists of a 512-dim FC layer, a Batch Normalization layer, a ReLU layer, a dropout layer, a $K$-dim FC layer ($K$ denotes the number of person identities) and a Cross Entropy layer. The person structure branch also contains similar layers. The only difference is that the dimension of fully connected layer is 256.

\subsection{Implementation Details}
\label{subsec:detail}
 The training images are augmented with horizontal flip and normalization and resized to $288\times 144$ and cropped to $256\times 128$ at the center with a small random perturbation. We use SGD with a mini-batch size of 32. We train RLD for 60 epochs. The learning rate starts from 0.1 and is decayed to 0.01 after 40 epochs. The backbone model is pre-trained on ImageNet and the learning rate for all the pre-trained layers are set to 0.1$\times$  of the base learning rate. We set ${\lambda}$ to 0.2. During testing, images are resized to $288\times 144$. We still extract global pooling features for feature representation.

\section{Experiment}
\label{sec:expe}
In this section, we evaluate our RLD method on three large-scale person ReID benchmark datasets, i.e., CUHK03, Market-1501 and DukeMTMC-reID, and present ablation studies to reveal the importance of each main component/factor of our method. We then reveal the benefits of the RLD model compared with state-of-the-art methods. We use CMC and mAP for evaluation. Note that, we use the same setting when comparing with baselines, including all of the hyper-parameters. And the baseline follows current works. That is, we simply set $\lambda$ to 0 in Eq. \ref{eq:loss}. To eliminate bias, we repeat the procedure 4 times to get an average performance for each experiment.

\textbf{\emph{Datasets.} }
The CUHK03 dataset contains 13,164 images of 1,467 identities. Each identity is observed by 2 cameras. It offers both hand-labeled and DPM-detected bounding boxes, and we use the latter in this paper. We adopt the new training/testing protocol proposed in \cite{Zhong2017CVPR}.

The Market-1501 dataset with six cameras is collected in Tsinghua University. Overlap exists among different cameras. Overall, this dataset contains 32,668 annotated bounding boxes of 1,501 identities. Among them, 12,936 images from 751 identities are used for training, and 19,732 images from 750 identities plus distractors are used for gallery. As for query, 3,368 hand-drawn bounding boxes from 750 identities are adopted. Each annotated identity is present in at least two cameras.

The DukeMTMC-reID dataset has 8 cameras. There are 1,404 identities appearing in more than two cameras and 408 identities (distractor ID) who appear in only one camera. Specially, 702 IDs are selected as the training set and the remaining 702 IDs are used as the testing set. In the testing set, one query image is picked for each ID in each camera and the remaining images are put in the gallery. In this way, there are 16,522 training images of 702 identities, 2,228 query images of the other 702 identities and 17,661 gallery images (702 ID + 408 distractor ID).

\subsection{Evaluation and Model Analysis}
To provide more insights on the performance of our approach, we conduct a lot of ablation studies by isolating each main component of our method.
\begin{figure*}[!t]
\centering
\subfloat[CUHK03.]{\includegraphics[width=2.1in]{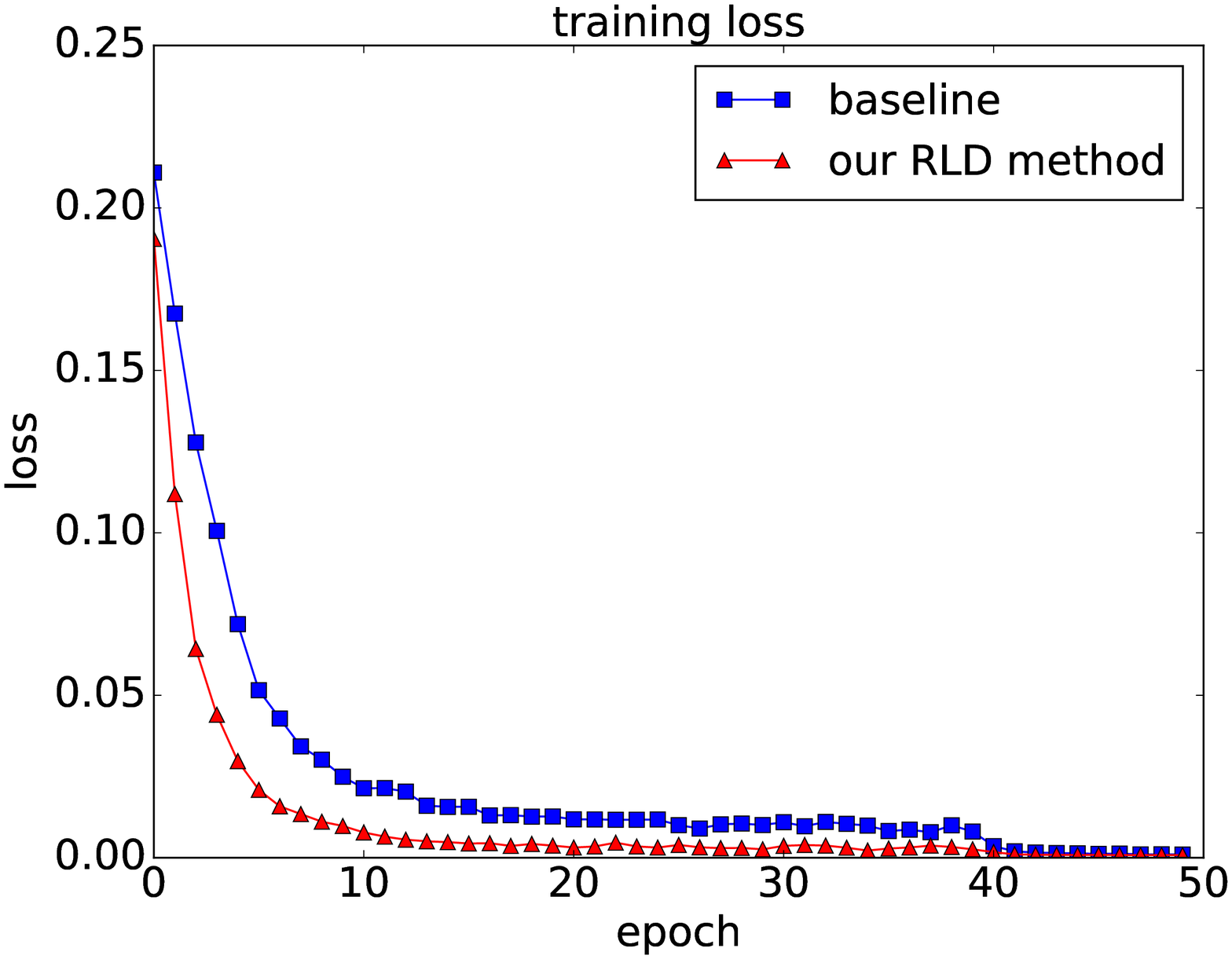}
\label{fig:stream}}
\hfil
\subfloat[Market-1501.]{\includegraphics[width=2.1in]{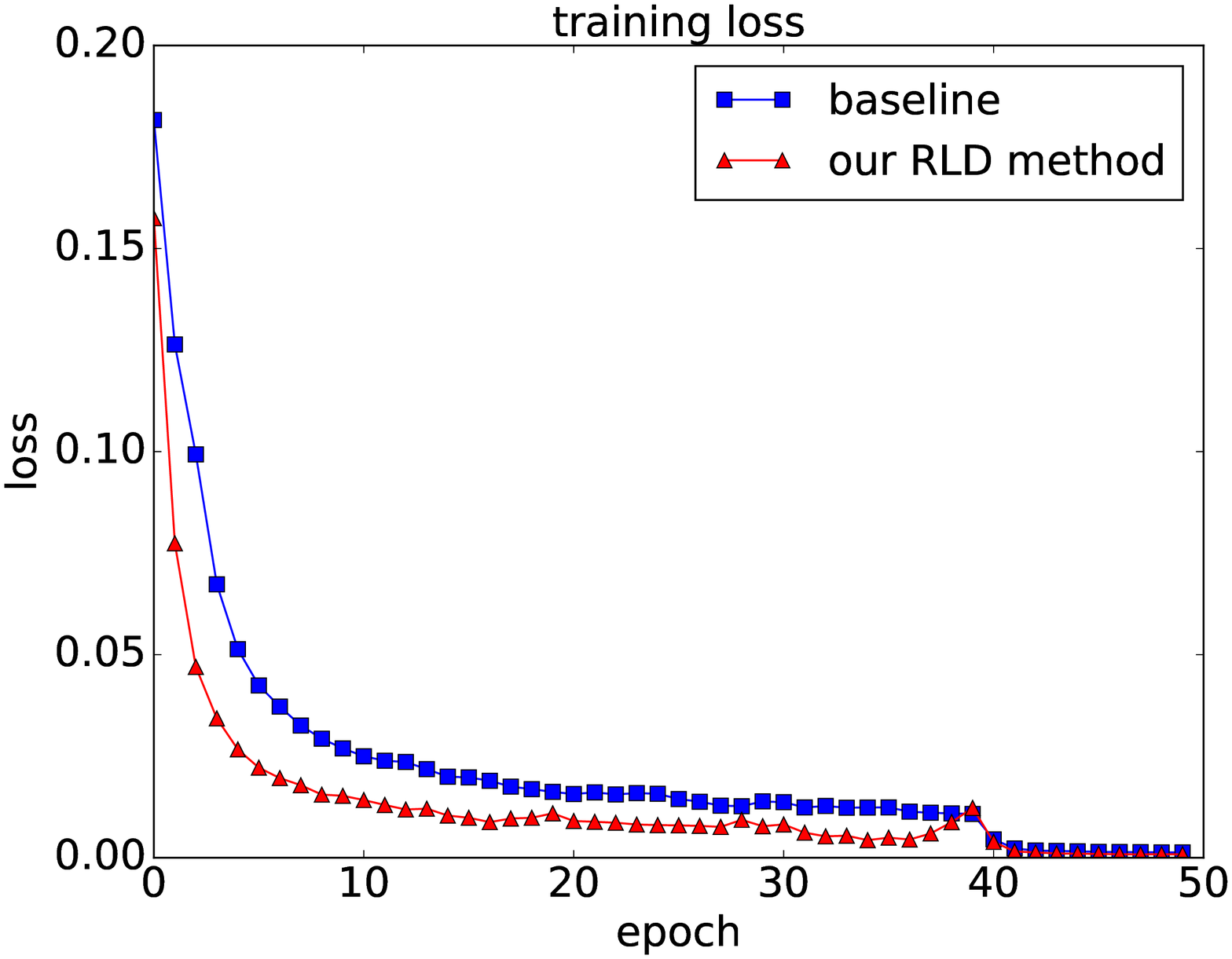}
\label{fig:joint}}
\hfil
\subfloat[DukeMTMC-reID.]{\includegraphics[width=2.1in]{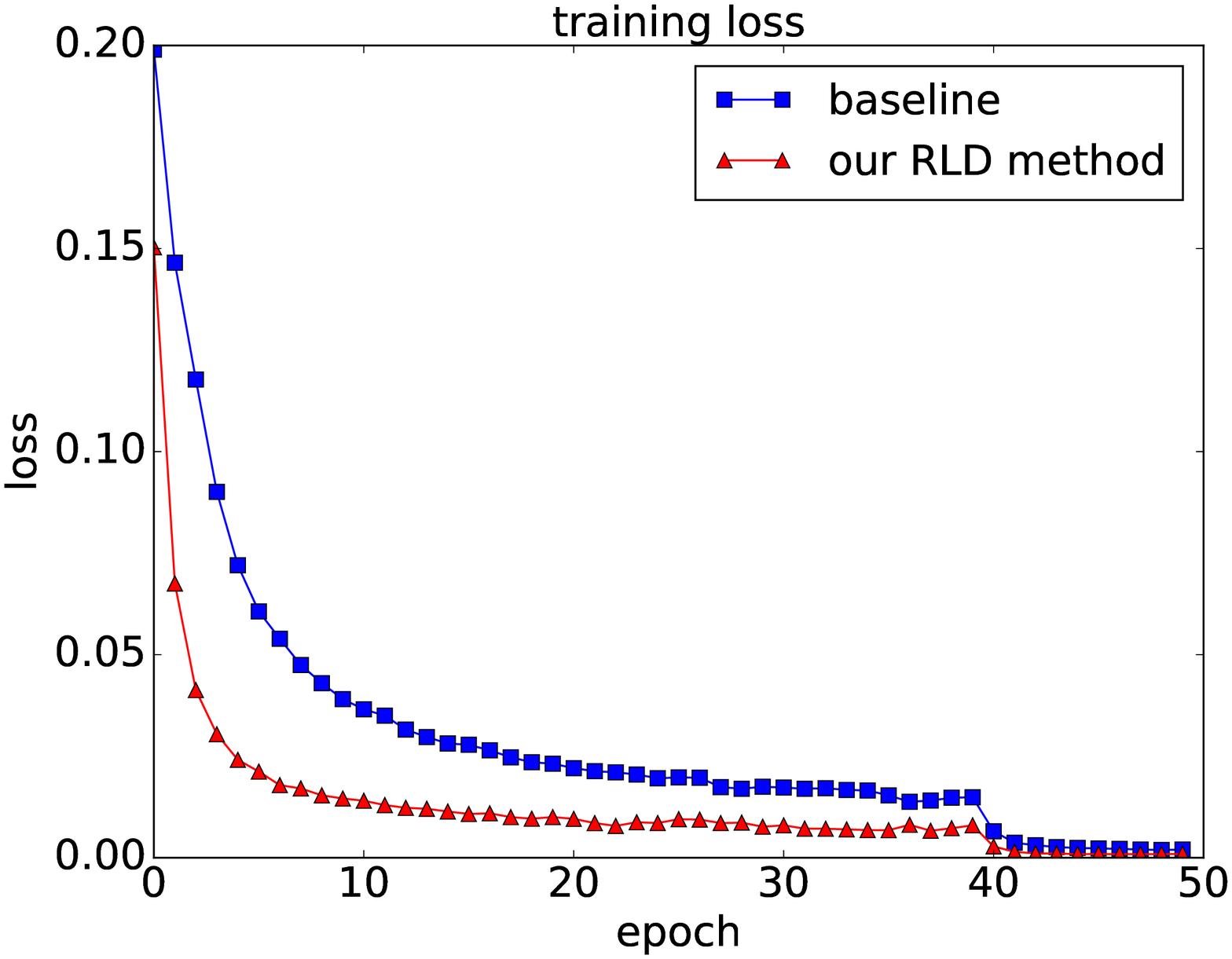}
\label{fig:fators1}}
\hfil
\caption{Comparisons of convergence.}
\label{fig:allresult}
\end{figure*}

\textbf{\emph{Effect of the person structure branch.}} To show the benefit of the person structure branch, we conduct an ablation study by isolating this constraint. We do this by removing the person structure branch and thus the framework is degraded as a baseline. In this experiment, we use ResNet-50 as the backbone architecture. As shown in Table \ref{tab:structure}, compared with the baseline, it is observed that the improvement in mAP is +5.5\%, +3.3\%, and +2.8\% on CUHK03, Market-1501 and DukeMTMC-reID, respectively. The improvement in rank-1 accuracy is +5.7\%, +1.8\%, and +1.8\%, respectively. Therefore, we conclude that the person structure branch works.
\begin{table*}
\begin{center}
\begin{tabular}{|c|c|c|c|c|c|c|c|c|c|c|c|c|}
\hline
\multirow{2}{*}{Methods} & \multicolumn{4}{c|}{CUHK03} & \multicolumn{4}{c|}{Market-1501} & \multicolumn{4}{c|}{DukeMTMC-reID} \\ \cline{2-13}
                         & r1    & r5    & r10  & mAP  & r1     & r5     & r10    & mAP   & r1   & r5   & r10  & mAP  \\ \hline
baseline                 & 46.8  & 67.5  & 76.5 & 43.0 & 86.6   & 94.8   & 96.6   & 68.0  & 75.9 & 87.3 & 90.7 & 56.9 \\ \hline
RLD (ours)                     & \textbf{52.5}  & \textbf{72.3}  & \textbf{79.8} & \textbf{48.5} & \textbf{88.4}   & \textbf{95.2}   & \textbf{96.9}   & \textbf{71.3}  &\textbf{ 77.7} & \textbf{88.4} & \textbf{91.4} & \textbf{59.7} \\ \hline
\end{tabular}
\end{center}
\caption{Effect of the person structure branch.}\label{tab:structure}
\end{table*}

\textbf{\emph{Comparison of convergence.}} To demonstrate the person structure constraint accelerates deep network training, we compare the RLD framework with the baseline that does not contain the person structure branch. We use ResNet-50 as the backbone architecture. The comparison of identity losses is shown in Figure \ref{fig:allresult}. We can see that our RLD obtains the faster convergence than the baseline on three datasets. Note that, the learning rate is decayed to 0.01 after 40 epochs and both identity losses are close to 0. We conclude that the person structure constraint guides the global feature representation learning.

\textbf{\emph{Effect of network architectures.}} To show the effect of our RLD method, we instantiate RLD with different network architectures, i.e., ResNet-18, ResNet-34, ResNet-50 and ResNet-101, as shown in Table \ref{tab:architectures}. We conduct four experiments on CUHK03. Compared with the baseline, we can see that our RLD method obtains 4.9\%, 3.3\%, 5.7\% and 3.5\% improvement in rank-1 accuracy on ResNet-18, ResNet-34, ResNet-50 and ResNet-101, respectively. The improvement in mAP is 5.2\%, 4.0\%, 5.5\% and 3.3\%, respectively. Therefore, we conclude that our RLD branch can be generalized to different network architectures.

\begin{table*}
\begin{center}
\begin{tabular}{|c|c|c|c|c|c|c|c|c|c|c|c|c|}
\hline
\multirow{2}{*}{Methods} & \multicolumn{3}{c|}{ResNet-18} & \multicolumn{3}{c|}{ResNet-34} & \multicolumn{3}{c|}{ResNet-50} & \multicolumn{3}{c|}{ResNet-101} \\ \cline{2-13}
                         & r1       & r5       & mAP      & r1       & r5       & mAP      & r1       & r5       & mAP      & r1        & r5       & mAP      \\ \hline
baseline                 & 39.9     &61.2     &36.3          & 46.0         &66.4          & 41.9         &46.8          & 67.5         &43.0          & 50.0       &71.3    &46.3          \\ \hline
RLD (ours)                      & \textbf{44.8}     & \textbf{66.0}     & \textbf{41.5}     & \textbf{49.3}     & \textbf{68.8}     & \textbf{45.9}     & \textbf{52.5}     & \textbf{72.3}     & \textbf{48.5}     & \textbf{53.5}      & \textbf{72.6}     & \textbf{49.6}     \\ \hline
\end{tabular}
\end{center}
\caption{Effect of network architectures.}\label{tab:architectures}
\end{table*}

\textbf{\emph{Parameters of the person structure branch.}} One may think if the better performance would be attributed to the increase of model parameters. To investigate this point, we compute the parameters of different components. We use ResNet-50 as the backbone architecture. As shown in Table \ref{tab:params}, we can see that the person structure branch is a small overhead, i.e., 0.8\% of entire model parameters. We conclude that the improvement of RLD is attributed to the benefit of the person structure constraint, but not the increase of model parameters.
\begin{table}
\begin{center}
\begin{tabular}{c|c|c|c}
\hline
         & baseline & person structure branch      &  $\Delta$\\ \hline
\#params & 25.0M & 0.2M &  0.8\%\\ \hline
\end{tabular}
\end{center}
\caption{Parameters of the person structure branch.}\label{tab:params}
\end{table}



\textbf{\emph{Influence of parameters $\lambda$.}} To investigate the impact of the important parameter ${\lambda}$ in Eq. \ref{eq:loss}, we conduct a sensitivity analysis experiment on CUHK03. We use ResNet-50 as the backbone architecture. As shown in Figure \ref{fig:allresult} (c) and (d), when ${\lambda}$ is in the range of 0.2$\sim$0.3, our model nearly keeps the best performance.
\begin{figure}
  \centering
  \includegraphics[width=0.45\textwidth]{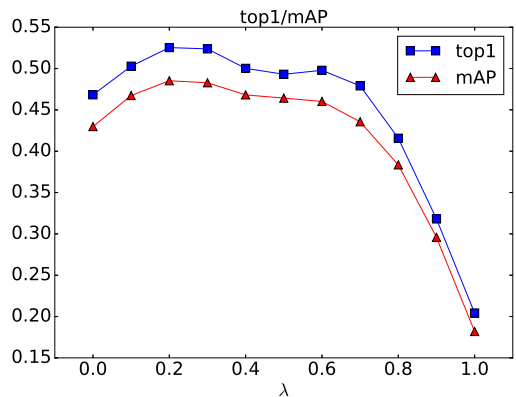}\\
  \caption{Influence of parameters $\lambda$. (Best viewed in color).}\label{fig:pooling}
\end{figure}

\begin{figure}
  \centering
  \includegraphics[width=0.45\textwidth]{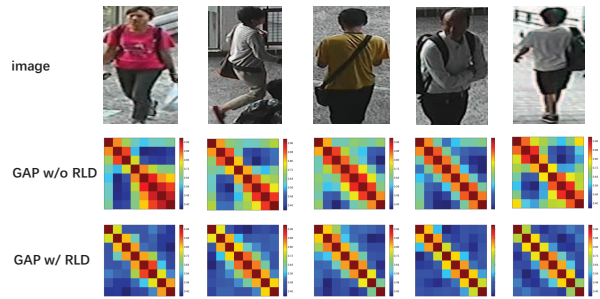}\\
  \caption{Comparison of visualization. Each column shows an image and its corresponding outputs with/without using the person structure branch for training.(Best viewed in color).}\label{fig:vis}
\end{figure}

\textbf{\emph{Comparison of Visualization.}} To show our RLD method can capture the person structure information, we \textbf{randomly select} 5 images and visualize the RLD matrixes with/without using the person structure branch for training, as shown in Figure \ref{fig:vis}. We can see that without using the person structure constraint, the strong dependencies of the RLD matrix elements are scattered. Differently, when using the person structure constraint, the strong dependencies are close to the diagonal line. This verifies the fact that local dependencies are often stronger than the remote ones for the person structure. We conclude that RLD can guide CNNs learn a structure-aware feature representation. Please refer to supplementary materials for more examples.

\subsection{Comparisons to the State-of-the-Art}
In this section, we compare our RDL with state-of-the-art methods.
\begin{table}
\begin{center}
\begin{tabular}{lcccc}
\hline
Methods      & R1   & R5   & R10  & mAP  \\ \hline
BoW+kissme \cite{zheng2015scalable}   & 44.4 & 63.9 & 72.2 & 20.8 \\
WARCA \cite{Jose2016eccv}       & 45.2 & 68.1 & 76.0 & -    \\
KLFDA \cite{karanam2016arXiv}       & 46.5 & 71.1 & 79.9 & -    \\ \hline
SOMAnet \cite{barbosa2018looking}     & 73.9 & -    & -    & 47.9 \\
SVDNet \cite{Sun_2017_ICCV}       & 82.3 & 92.3 & 95.2 & 62.1 \\
PAN  \cite{zheng2017axkiv}        & 82.8 & -    & -    & 63.4 \\
Transfer \cite{geng2016deep}    & 83.7 & -    & -    & 65.5 \\
Triplet Loss \cite{HermansBL17} & 84.9 & 94.2 & -    & 69.1 \\
DML \cite{zhang2017deep}          & 87.7 & -    & -    & 68.8 \\ \hline
MultiRegion \cite{ustinova2017multi}  & 66.4 & 85.0 & 90.2 & 41.2 \\
HydraPlus \cite{liu2017hydraplus}    & 76.9 & 91.3 & 94.5 & -    \\
PAR \cite{zhao2017deeply}          & 81.0 & 92.0 & 94.7 & -    \\
MultiLoss \cite{li2017person}    & 83.9 & -    & -    & 64.4 \\
PDC* \cite{Su2017ICCV}         & 84.4 & 92.7 & 94.9 & 63.4 \\
PartLoss \cite{yao2017deep}    & 88.2 & -    & -    & 69.3 \\
\hline
RLD (ours)            & 88.4 & 95.2 & 96.9 & 71.3 \\
RLD+Era \cite{random_erase}      & \textbf{90.2} & \textbf{96.4} & \textbf{97.8} & \textbf{76.1} \\
\hline
\end{tabular}
\end{center}
\caption{Comparison of the proposed method with the art on Market-1501. The compared methods are categorized into 3 groups. Group 1: hand-crafted methods. Group 2: deep learning methods employing global feature. Group 3: deep learning methods employing part features. * denotes those requiring extra part labels. Our method is denoted by RLD.}\label{tab:market}
\end{table}

\begin{table}
\begin{center}
\begin{tabular}{l|cccc}
\hline
\multirow{2}{*}{Methods} & \multicolumn{2}{c}{CUHK03} & \multicolumn{2}{c}{Duke.} \\ \cline{2-5}
                         & rank-1       & mAP       & rank-1        & mAP        \\ \hline
BoW+kissme \cite{zheng2015scalable}               & 6.4           & 6.4      & 25.1         & 12.2              \\
LOMO+XQDA \cite{liao2015person}               & 12.8          & 11.5     & 30.8         & 17.0             \\
GAN \cite{zheng2017unlabeled}                     & -             & -        & 67.7         & 47.1                \\
PAN \cite{zheng2017axkiv}                      & 36.3          & 34.0     & 71.6         & 51.5            \\
SVDNet \cite{Sun_2017_ICCV}                  & 41.5          & 37.3     & 76.7         & 56.8             \\
MultiScale \cite{chen2017cvprw}              & 40.7          & 37.0     & 79.2         & 60.6             \\
TriNet+Era \cite{random_erase}              & 55.5          & 50.7     & 73.0         & 56.6             \\
SVDNet+Era \cite{random_erase}              & 48.7          & 43.5     & 79.3         & 62.4             \\
\hline
RLD (ours)                      & 52.5          & 48.5     & 77.7         & 59.7             \\
RLD + Era \cite{random_erase}                  & \textbf{56.2}          & \textbf{52.2}     & \textbf{79.5}         & \textbf{63.4}             \\
\hline
\end{tabular}
\end{center}
\caption{Comparison with prior art on DukeMTMC-reID and CUHK03. Rank-1 accuracy (\%) and mAP (\%) are shown..}\label{tab:cuhk03}
\end{table}

\textbf{\emph{Evaluations on Market-1501.}} We compare the proposed RLD and RLD+Era with fifteen existing state-of-the-art methods, which can be grouped into three categories, i.e., handcrafted feature methods, deep learning methods with global features, and deep learning methods with part features. Among them, PDC requires extra part labeling to align parts. Differently, our RLD can learn a person structure-aware feature representation without using extra annotation. Comparisons on Market-1501 are shown in Table \ref{tab:market}.


\textbf{\emph{Evaluations on CUHK03 and DukeMTMC-reID.}} We compare our RLD and RLD+Era methods with eight state-of-the-art methods on the CUHK03 and DukeMTMC-reID (Duke. for short) datasets , respectively. As shown in Table \ref{tab:cuhk03}, it is encouraging to see that our approach significantly outperforms the competing methods. For example, RLD+Era obtains a good performance on both CUHK03 (56.2\% rank-1 and 52.2\%mAP) and DukeMTMC-reID (79.5\% rank-1 and 63.4\% mAP), while SVDNet+Era or TriNet+Era only obtains a good performance either on CUHK03 or DukeMTMC-reID.


\textbf{\emph{Remarks.}}  The RLD method is easily implemented. Without bells and whistles, our RLD model achieves competitive results with state-of-the-arts. While outside the scope of this work, we expect many such techniques (e.g., $1\times 1$ last convolutional kernel, $384\times 128$ image size, multi-loss or re-ranking) to be applicable to ours.


\section{Conclusion}
\label{sec:conc}
In this paper, we propose a novel Relative Local Distance (RLD) method to model the underlying person structure for person re-identification (re-ID). To this a relative local distance matrix is introduced to learn a structure-aware feature representation. With the discovered underlying person structure, the RLD method builds a bridge between the global and local feature and thus improves the capacity of feature representation for person re-ID. In summary, our contribution is three-fold. First, it is the first time the relative local distance is proposed to exploit the underlying person structure without using any pose annotation for person re-ID. Second, with the joint training of the person identity loss and person structure loss, RLD significantly accelerates deep network training and improves the performance of person re-ID. Third, the experimental results show the effectiveness of RLD on three benchmark datasets.

We intend to extend this work in two directions. First, we intend to generalize the relative local distance model to general object recognition in an end-to-end way. We also hope this insight can be further developed in the fine-grained classification problem. Second, we intend to study how to further exploit discriminative features from the last convolutional layers instead of using global average pooling.

{\small
\bibliographystyle{ieee}
\bibliography{egbib}
}

\end{document}